\documentclass[10pt,journal]{IEEEtran}
\usepackage{amsmath}
\usepackage{amsfonts}
\usepackage{amssymb}
\usepackage{array}
\usepackage{multirow}
\usepackage{multicol}
\usepackage{graphicx}
\usepackage{subfig}
\usepackage{threeparttable}
\usepackage{color}
\usepackage{soul}

\usepackage{algorithm}
\usepackage[noend]{algpseudocode}

\begin{document}

\title{Causal inference of brain connectivity from fMRI with $\psi$-Learning Incorporated Linear non-Gaussian Acyclic Model ($\psi$-LiNGAM)}

\author{Aiying~Zhang, Gemeng~Zhang, Biao~Cai, Wenxing~Hu, Li~Xiao,
        Tony~W.~Wilson,~\IEEEmembership{Senior Member,~IEEE},
        ~Julia~M.~Stephen,~\IEEEmembership{Senior Member,~IEEE},
        ~Vince~D.~Calhoun,~\IEEEmembership{Fellow,~IEEE，}
        and Yu-Ping~Wang,~\IEEEmembership{Senior Member,~IEEE}

\thanks{A. Zhang, G. Zhang, B. Cai, W. Hu, L. Xiao and Y.-P. Wang are with the Department of Biomedical Engineering, Tulane University, New Orleans, LA 70118, USA (e-mail: wyp@tulane.edu).}
\thanks{T. W. Wilson is with the Department of Neurological Sciences, University
of Nebraska Medical Center, Omaha, NE 68198 USA.}
\thanks{J. M. Stephen and V. D. Calhoun are with Mind Research Network,
Albuquerque, NM 87106 USA.}
\thanks{V. D. Calhoun is with Tri-institutional Center for Translational Research in Neuroimaging and Data Science (TReNDS), Georgia State University, Georgia Institute of Technology, Emory University, Atlanta, GA, 30303 USA.}
}

\maketitle

\begin{abstract}

Functional connectivity (FC) has become a primary means of understanding brain functions by identifying brain network interactions and, ultimately, how those interactions produce cognitions. A popular definition of FC is by statistical associations between measured brain regions. However, this could be problematic since the associations can only provide spatial connections but not causal interactions among regions of interests. Hence, it is necessary to study their causal relationship. Directed acyclic graph (DAG) models have been applied in recent FC studies but often encountered problems such as limited sample sizes and large number of variables (namely high-dimensional problems), which lead to both computational difficulty and convergence issues. As a result, the use of DAG models is problematic, where the identification of DAG models in general is nondeterministic polynomial time hard (NP-hard). To this end, we propose a $\psi$-learning incorporated linear non-Gaussian acyclic model ($\psi$-LiNGAM). We use the association model ($\psi$-learning) to facilitate causal inferences and the model works well especially for high-dimensional cases. Our simulation results demonstrate that the proposed method is more robust and accurate than several existing ones in detecting graph structure and direction. We then applied it to the resting state fMRI (rsfMRI) data obtained from the publicly available Philadelphia Neurodevelopmental Cohort (PNC) to study the cognitive variance, which includes 855 individuals aged 8–22 years. Therein, we have identified three types of hub structure: the in-hub, out-hub and sum-hub, which correspond to the centers of receiving, sending and relaying information, respectively. We also detected $16$ most important pairs of causal flows. Several of the results have been verified to be biologically significant. To facilitate reproducible research, we make the code to be publicly available at https://github.com/Aiying0512/psi-LiNGAM.
\end{abstract}

\begin{IEEEkeywords}
Causality, Bayesian networks, fMRI, LiNGAM, Brain functional connectivity。
\end{IEEEkeywords}

\section{Introduction}

Functional magnetic resonance imaging (fMRI) has been commonly used to noninvasively detect human brain activity by measuring the blood oxygenation-level dependent (BOLD) changes. In the past decades, a large number of models have been developed to analyze brain functional connectivity (FC) using fMRI. By incorporating graph theory, the brain can be described as a graph, where a node represents a well-defined region of interest (ROI) and an edge represents a functional interconnection between the nodes (ROIs) it connects. Then, the brain FC can be mathematically estimated and quantified. Based on the representation form of the graph, methods for FC estimation can be roughly divided into two categories: undirected graphical models, and directed acyclic graph (DAG) models. The undirected graphical models characterize the brain FC through the association coefficients among various ROIs, which reflect their spatial relationships. For instance, the Pearson correlation is a typical measurement in undirected graphical models from static to dynamic FC estimations \cite{Cai2018, zgm2019}, which describes the linear correlation between the ROIs. In a complex system like the brain, the Pearson correlation is much weaker marginally \cite{Liang2015}. That is, all nodes (ROIs) are directly or indirectly correlated and it is difficult to distinguish significant connections through a dense network constructed by Pearson correlations. To this end, partial correlations have been proposed to explore direct associations between two nodes, removing the effect of other random variables \cite{Zhang2018,zhangTMI2019}. Approaches to characterize statistical associations are often used for estimating brain network interactions, but fail to reveal the direction of information flow behind these interactions. It is natural to shift the focus of FC from association to causal interactions for in-depth research \cite{reid2019}.

Directed acyclic graph (DAG) models, also known as Bayesian networks, are designed to model causal relationships in complex systems. Causality models pinpoint the key connectivity characteristics and simultaneously remove some redundant features for diagnosis. Methods for DAG identification can be divided into four categories: constraint-based methods, score-based methods, non-Gaussian based methods, and hybrids of these categories. Constraint-based methods, such as the prominent PC algorithm \cite{pc2010}, first learn the skeleton of the DAG from conditional independence relationships and then orient the edges. Score-based methods, on the other hand, posit a scoring criterion for each possible DAG model, and then search for the graph with the highest score given the observations. An example is the greedy equivalence search (GES) algorithm proposed by Chickering \cite{GES}, which greedily optimizes the $l_0$-penalized likelihood such as the Bayesian Information Criterion (BIC). Both the constraint-based and the score-based methods are not optimal for predicting the direction of causal relationships but are accurate in identifying the causal skeleton (graph structure without directions) in fMRI study \cite{Smith2011}. The non-Gaussian based methods, which refer to the Linear Non-Gaussian Acyclic Model (LiNGAM), aim to estimate linear Bayesian networks for continuous data using the non-Gaussian information. The key aspect of LiNGAM is the use of non-Gaussian data, which makes it possible to identify more of the graph structure than the traditional Gaussian setting. Several methods have been proposed in the literature, such as ICA-LiNGAM \cite{Shimizu2006}, direct LiNGAM \cite{Shimizu2011}, and pairwise LiNGAM \cite{Smith2013}. However, compared to the other two categories of methods, LiNGAM requires a larger number of data points in the relevant dimension to converge to the true graph.

In biomedical applications, we are often faced with small sample size problems where the number of variables/nodes greatly exceeds the number of samples/observations, i.e., high dimensional cases. As a result, both computational difficulty and convergence issue often arise based only on the observations. This is especially the case for the estimation of DAG models, which is in general nondeterministic polynomial time hard (NP-hard) \cite{Heckerman1995}. To this end, we propose a $\psi$-learning incorporated linear non-Gaussian acyclic model ($\psi$-LiNGAM), particularly for high-dimensional cases. The $\psi$-learning method \cite{Liang2015} was first proposed to fast estimate the undirected graph with the equivalent partial correlations,i.e., the $\psi$-correlation. The method works well especially for high-dimensional cases, with successful applications to gene regulatory network inference \cite{jia2017learning} and  brain FC analysis by us recently \cite{Zhang2018}. Herein, we apply the $\psi$-learning method to refine the set of likely causal connectivity, facilitating causal inferences with fast computation. Following the same idea of LiNGAM (e.g., \cite{Shimizu2011}), we assume that the observations are non-Gaussian and continuous, and incorporate prior information from direct LiNGAM into the DAG estimation. To acquire the prior knowledge, we estimate the undirected graph first, since the skeleton of the directed graph is always included in it. To be more specific, we apply the nonparanormal transformation \cite{Liu2009} first, which converts the problem to a Gaussian graphical model (GGM), and then adopt the $\psi$-learning  to identify the undirected graph structure. The contributions of this paper are generally two-folds. Mathematically, we overcome the high-dimensionality difficulty of the LiNGAM methods and increase the convergency speed to the true DAG, even under the condition of small sample size. Biomedically, we apply the proposed model to the resting-state fMRI (rsfMRI) data from PNC, aiming to explain the cognitive variance through the directed functional connectivity differences. Unlike previous studies using association models \cite{Cai2018,Zhang2018}, our work is able to orient the causal directions and identify three types of hub structures: in-hub, out-hub and sum-hub from the directed graphs.

The remainder of this paper is organized as follows. In Section \uppercase\expandafter{\romannumeral2}, we first introduce some background knowledge of directed graphs and then describe the proposed $\psi$-LiNGAM method. Both simulation studies and the rsfMRI analysis using PNC data are shown in Section \uppercase\expandafter{\romannumeral3}, followed by some discussions and concluding remarks in the last two sections.

\section{Methods}

In this section, we first introduce some basic concepts of DAGs and the relationship between directed and undirected graphs in Section \uppercase\expandafter{\romannumeral2}.A. Then we describe the LiNGAM and the method for prior knowledge estimation in Section \uppercase\expandafter{\romannumeral2}.B and \uppercase\expandafter{\romannumeral2}.C, respectively. Finally, we summarize the $\psi$-LiNGAM algorithm in Section \uppercase\expandafter{\romannumeral2}.D.

\subsection{Background: directed graph}

\begin{figure*}[h!]
\centering
\includegraphics[width=5.5in]{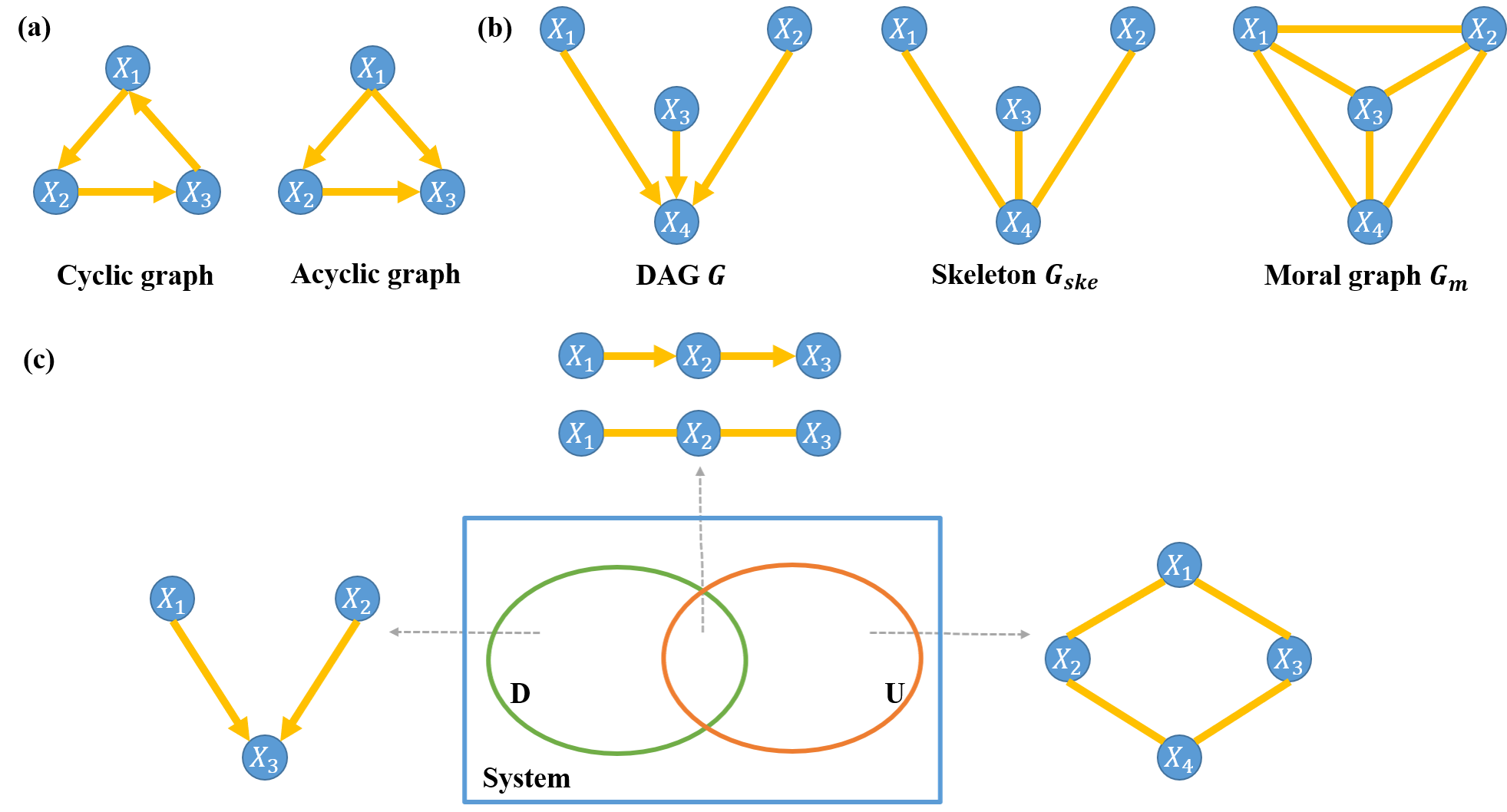}
\caption{Illustrations of the concepts in directed graphs. (a) gives an example of the cyclic and acyclic graphs. (b) shows the relationships between a DAG $G$ and its corresponding skeleton $G_{ske}$ and moral graph $G_m$. (c) is the Venn diagram describing the relationships in a network system and the representations of the directed and undirected graphs of the system. D represents the directed graph and U represents the undirected graph.}\label{fig:1}
\end{figure*}

A graph $G = (V, E)$ is composed of a node set $V = \{1,2,...,p\}$ and an edge set $E \subset V \times V$. In our setting, the nodes in set $V$ correspond to the components of a random vector $\mathbf{X} = (X_1, X_2,..., X_p)^T$. An edge $(i,j)$ is directed if $(i,j) \in E$ but $(j,i) \notin E$, and we denote it as $i \to j$ and node $i$ is called a parent of node $j$. A directed acyclic graph (DAG) $G$ means all the edges in $G$ are directed and there is no circle in graph $G$ (see Fig. \ref{fig:1} (a)). An edge $(i,j)$ is undirected if $(i,j) \in E$ and $(j,i) \in E$, and we denote it as $i - j$. An undirected graph $G_{und}$ is composed of undirected edges. The skeleton of a DAG $G$, $G_{ske}$, is the undirected graph obtained from $G$ by substituting undirected edges for directed edges (see Fig. \ref{fig:1} (b)).

The two graphical structures, directed and undirected graphs, express different independence properties in a system. Fig. \ref{fig:1} (c) gives an illustration of a directed/ undirected graph whose independence properties cannot be expressed by each other. A directed graph can be converted into an undirected graph, called a moral graph \cite{bishop2006pattern}. In general, we first add additional undirected links between all pairs of parents for each node in the graph and then drop the arrows on the original links to give the moral graph $G_m$. The relationship among the skeleton of the DAG $G_{ske}$, the moral graph $G_m$ and the undirected graph $G_{und}$ is $G_{ske} \subset G_m \subset G_{und}$.

\subsection{LiNGAM}

The linear non-Gaussian acyclic model (LiNGAM) was first proposed by Shimizu et al. \cite{Shimizu2006}, which is a Bayesian network (BN) using a structural equation model (SEM) for non-Gaussian variants. LiNGAM assumes that the casual relationships of the variables can be represented graphically by a DAG $G = (V, E)$, where the node set $V = \{1,2,...,p\}$ represents the corresponding variables and $E \subset V \times V$ denotes the directed edges. Let $\bold{B} =\{b_{ij}\} \in \mathbb{R}^{p \times p}$ be the weighted adjacency matrix specifying the edge weights of the underlying DAG $G$. The observed random vector $\bold{X} = (X_1, X_2, \dots, X_p)^T \in \mathbb{R}^p$ is assumed to be generated from the following linear SEM,
\[
\bold{X} = \bold{B}^T \bold{X} + \boldsymbol{\epsilon},
\]
where $\boldsymbol{\epsilon} = (\epsilon_1, \epsilon_2, ..., \epsilon_p)^T$ is a continuous random vector; the $\epsilon_i$'s, $\forall i =1,2,..., p$ have non-Gaussian distributions with non-zero variances, and are independent of each other.

A property of acyclicity is that there exists at least one permutation $\pi$ of $p$ variables such that $b_{ij} = 0$, $\forall \pi(i) < \pi(j)$. In other words, the weight matrix $\bold{B}$ can be reordered to a strictly lower triangular matrix according to the permutation $\pi$. The goal of LiNGAM is to find the correct permutation and estimate the weight matrix $\bold{B}$. Since the components of $\boldsymbol{\epsilon}$ are independent and non-Gaussian, Shimizu et al. \cite{Shimizu2006} first proposed the independent component analysis (ICA) based algorithm known as ICA-LiNGAM. However, like most ICA algorithms, the convergency of ICA-LiNGAM is affected by the initial state, and an appropriate parameter selection is not straightforward. The direct LiNGAM \cite{Shimizu2011} method has been developed without the need for initial guess or algorithmic parameters, which also has the guaranteed convergence. However, it requires a large number of samples to reach the true graph and the computational time is much longer than the ICA-LiNGAM. Fortunately, the direct LiNGAM is flexible to incorporate prior knowledge. With a proper selection of priors, both the convergence and the computation can be largely accelerated due to the significant shrinkage of searching space.

\subsection{Prior knowledge estimation}

We have discussed the relationship between directed graphs and undirected graphs in Section \uppercase\expandafter{\romannumeral2}.A. The undirected graph can be treated as the upper bound of the DAG, which can be estimated first as the prior information. Our goal in this step is to eliminate as many confounding edges as possible and only keep the direct associated edges, and therefore we select partial correlation coefficients as the measure of dependencies. Gaussian graphical models (GGMs) have become a popular tool for the estimation of undirected graphs using partial correlation coefficients due to their mathematical simplicity. To be more specific, let $\bold{X}=(X_1, X_2, \dots, X_p)^T$ denote a $p$-dimensional random vector following a multivariate Gaussian distribution $N(\boldsymbol{\mu},\boldsymbol{\varSigma})$, where $\boldsymbol{\mu}$ and $\boldsymbol{\varSigma}$ denote the unknown mean and the covariance matrix, respectively. It has been proven that the partial correlation between $X_i$ and $X_j$ on the condition of all other variables can be expressed as
\begin{equation}
 \rho_{ij|V\backslash \{ij\}} = - \frac{\varOmega_{ij}}{\sqrt{\varOmega_{ii}\varOmega_{jj}}}, i,j = 1,...,p,
\end{equation}
where $\boldsymbol{\varOmega}$ is the precision matrix (i.e., the inverse of the covariance matrix $\boldsymbol{\varSigma}$), $\varOmega_{ij}$ denotes the ($i$,$j$)th entry of $\boldsymbol{\varOmega}$  and $V = \{1,2,...,p\}$ denotes the index set of variables \cite{Dempster1972}, \cite{Lauritzen1996}. Thus, the construction of undirected graphs using GGMs is equivalent to the estimation of the precision matrices.

In our model, the variables are assumed to be non-Gaussian continuous, thus, the original GGMs with Gaussian assumption cannot be directly applied. To this end, we adopt the nonparanormal transformation proposed by Liu et al. \cite{Liu2009}, known as a semiparametric Gaussian copula model, to render data normal. Subsequently, we adopt the $\psi$-learning method \cite{Liang2015} to infer the GGM structure due to its computational efficiency and flexible framework as demonstrated by our previous work \cite{Zhang2018}.

\subsection{$\psi$-LiNGAM}

We now illustrate the $\psi$-LiNGAM we proposed for high dimensional DAG estimation. First, we follow the same definition of the prior matrix $\bold{A}^{prior} = \{a^{prior}_{ij}\}$ as in \cite{Shimizu2011}, which is given as follows:
\begin{equation}
a^{prior}_{ij}:=
\left\{
\begin{array}{lr}
 0, \text{if there is no directed edge from i to j}  \\

 1, \text{if there is a directed edge from i to j}  \\
-1, \text{the edge status is not sure.}\\
\end{array}
\right.
\end{equation}

The procedure of the $\psi$-LiNGAM algorithm is summarized in Algorithm \ref{latent}. The nonparanormal transformation can be implemented through the R package \emph{huge}. The R package \emph{equSA} is designed for the $\psi$-learning method. Since we intend to find the upper bound of the DAG, we set a broader significance level with $\alpha_2 = 0.2$. The code for direct LiNGAM is publicly available at https://sites.google.com/site/sshimizu06/lingam.

\algsetblock[Name]{Start}{End}{3}{0.2cm}
\renewcommand{\algorithmicrequire}{\textbf{Input:}}
\renewcommand{\algorithmicensure}{\textbf{Output:}}
\renewcommand{\algorithmicprocedure}{\textbf{Procedure:}}
\begin{algorithm}
\caption{$\psi$-LiNGAM algorithm}\label{latent}
\begin{algorithmic}
\Require Observed $n$ sample vectors $\bold{x} = (\bold{x_1}, \bold{x_2},..., \bold{x_p}) \in \mathbb{R}^{n \times p}$, where the $\bold{x_i}$'s are non-Gaussian continuous.
\Ensure Estimated weighted DAG adjacency matrix $\hat{\bold{B}}$
\State 1. Prior knowledge estimation: undirected graph structure estimation.
\Start
\State a. Use the nonparanormal transformation \cite{Liu2009} to render $\bold{x}$ normal (Gaussian).
\State b. Apply $\psi$-learning method \cite{Liang2015} to infer the GGM structure and acquire the edge set $E^{prior}$ of the undirected graph.
\State c. Extract the prior matrix $\bold{A}^{prior}$ from $E^{prior}$, where $a_{ij}^{prior} = -1$, if $(i,j) \in E^{prior}$ and otherwise $a_{ij}^{prior} = 0$.
\End
\State 2. Identify the casual order $\pi$ using the direct LiNGAM with the prior matrix $\bold{A}^{prior}$ \cite{Shimizu2011}.
\State 3. Obtain the estimated weighted DAG adjacency matrix $\hat{\bold{B}}$.
\Start
\State a. Construct a strictly lower triangular matrix $\tilde{\bold{B}}$ by following the causal order $\pi$, and the corresponding $\tilde{\bold{A}}^{prior}$ with the same order.
\State b. Estimate the connection strengths $\tilde{\bold{b}}_{j}^T = (\tilde{b}_{1j}, \tilde{b}_{2j}, ..., \tilde{b}_{pj} )$ consistent with $\tilde{\bold{A}}^{prior}$ by solving sparse regressions of the form
\[
\hat{\tilde{\bold{b}}}_j = \arg \min \limits_{supp(\tilde{\bold{b}}_j) \subset supp(\tilde{\bold{a}}_j^{prior})} ||\bold{x_j}-\bold{x}\tilde{\bold{b}}_j||_2^2
\]
\State c. Obtain $\hat{\bold{B}}$ by converting $\tilde{\bold{B}}$ to the original order.

\End

\end{algorithmic}
\end{algorithm}

\section{Results}

\subsection{Simulation studies}

In this section, we evaluated our proposed model through a series of simulation studies. We simulated the random DAG $G$ through the R package \emph{pcalg} with the edge probability $d/(p-1)$, where $d$ is an edge degree parameter and $p$ is the total number of variables. Given $G$, we assigned uniformly random weights to the edges to obtain the weighted adjacency matrix $\bold{B}$: $b_{ij} \sim \rm Unif (-0.8,-0.3)\cup (0.3,0.8)$, if $\ b_{ij} \in E$; otherwise $b_{ij}=0$. Given $\bold{B}$, we generated $\bold{x} = \bold{B}^T \bold{x} + \boldsymbol{\epsilon} \in \mathbb{R}^p$ from two non-Gaussian noise selections: Exponential (Exp) and Chi-squared (Chisq) noises. The exponential noise is set to have rate $1$, i.e., $\epsilon_i \sim exp(1)$, $i=1,2,...,p$, and the chi-squared noise is set to be central with degree of freedom $1$, i.e. $\epsilon_i \sim \chi^2_1$, $i=1,2,...,p$. We illustrated the performance of the proposed $\psi$-LiNGAM under the cases of small sample size and high dimensionality. We set $n=100$ and sampled the random vectors $\bold{x} \in \mathbb{R}^{ n \times p}$ for each noise selection with $p = 50, 100, 200$ and the degree parameter $d =1, 2, 4$. For each scenario, we simulated 10 datasets independently. We assessed the performance of the four methods through the true positive rate (TPR), false discovery rate (FDR), and structural hamming distance (SHD) \cite{SHD2006}. SHD is a commonly used metric based on the number of operations needed to transform the estimated DAG into the true graph \cite{Kalisch2007}. In simple terms, SHD counts the total number of edge insertions, deletions or flips in the transformation.

%
%
%

\begin{figure*}
\centering
\subfloat[Subfigure 1 list of figures text][Exp]{\includegraphics[width = 6in]{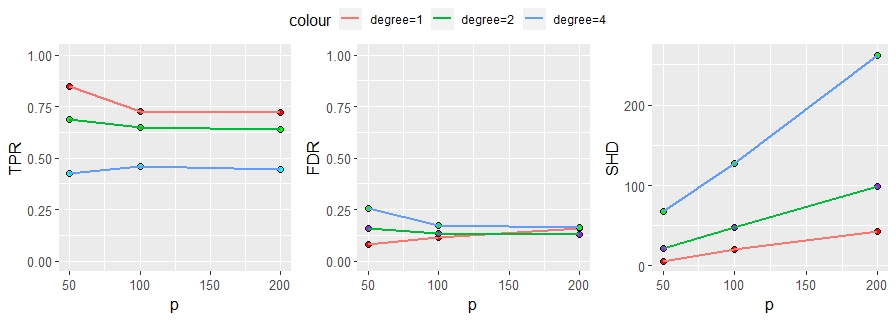}}
\qquad
\subfloat[Subfigure 1 list of figures text][Chisq]{\includegraphics[width = 6in]{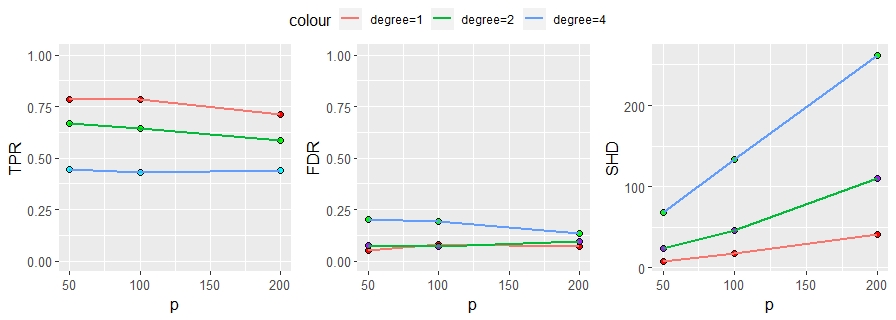}}
\qquad
\caption{Simulation results under small sample size and high-dimensional cases, which represent the average performance in terms of TPR, FDR and SHD under various variable ($p=50,100,150$), degree parameter ($d=1,2,4$) and noise (Exp, Chisq) settings with sample size $n=100$.}\label{fig:2}
\end{figure*}

Fig. \ref{fig:2} shows the performance of $\psi$-LiNGAM under small sample size and high-dimensional settings with $n=100$ (i.e., $n < p$). As the graph gets denser ($d$ increases) or larger ($p$ increases),  although the TPR descends the FDR remains at a low range. To further demonstrate the superiorty of the $\psi$-LiNGAM, we compared it with $4$ existing methods, which were the PC algorithm \cite{pc2010}, GES \cite{GES}, the ICA-LiNGAM \cite{Shimizu2006} and the direct LiNGAM \cite{Shimizu2011} with $n=500$ (i.e., $n > p$) in \textbf{Appendix.A}. (The two LiNGAM methods cannot be applied for high-dimensional cases.)
From the results between $\psi$-LiNGAM and direct LiNGAM, $\psi$-LiNGAM has significantly solved the convergence speed issue. In fact, the $\psi$-LiNGAM has outperformed the other methods under each setting, especially under large variable numbers and/or low degree parameter settings (see \textbf{Appendix.A} Figure A, B). Compared to direct LiNGAM, the computational complexity of $\psi$-LiNGAM is also favorable. The mean CPU times of the two methods under various variable sizes $p$'s on a 4 GHz computer are shown in TABLE \uppercase\expandafter{\romannumeral1}.

\begin{table}[h!]
\centering
\caption{The mean CPU times (in seconds) for $d=1, 2, 4$ under various variable sizes $p$'s. }
\begin{tabular}{|c|c|c|c|}
\hline
Method & $p=50$ & $p=100$ & $p=200$ \tabularnewline
\hline
$\psi$-LiNGAM & $28.86$ & $291.40$ & $2130.26$  \tabularnewline
Direct-LiNGAM & $51.83$ & $499.00$ & $3806.14$  \tabularnewline
\hline
\end{tabular}
\end{table}

\subsection{Brain network analysis based on fMRI}

\subsubsection{Data description}
In this work, we intend to investigate the relationship between the variance of cognitive abilities and the brain functional activity. The dataset we used is the publicly available Philadelphia Neurodevelopmental Cohort (PNC), which is a large-scale collaborative study of the Brain Behavior Laboratory at the University of Pennsylvania and the Children’s Hospital of Philadelphia \cite{Satt2016}. The data contains (among other fMRI modalities \cite{Satt2014}) rsfMRI for nearly $900$ subjects aged from $8$ to $21$ years old. All MRI scans were acquired on a single 3T Siemens TIM Trio whole-body scanner. Blood oxygenation level-dependent (BOLD) fMRI was acquired using a whole-brain, single-shot, multislice, gradient-echo (GE) echoplanar (EPI) sequence of $124$ volumes (372s) with the following parameters TR/TE=3000/32 ms, ﬂip $=90^\circ$, FOV$=192 \times 192$ mm, matrix $= 64 \times 64$, slice thickness/gap $=3 mm/0 mm$. The resulting nominal voxel size was $3.0 \times 3.0 \times 3.0 $mm \cite{Satt2014}. Standard preprocessing steps were applied using SPM12 (https://www.fil.ion.ucl.ac.uk/spm/software/spm12/), including motion correction, spatial normalization to standard Montreal Neurological Institute (MNI) space, and spatial smoothing with a $3$ mm full width at half max (FWHM) Gaussian kernel. The effect of head motion ($6$ parameters) was further regressed out and BOLD time series went through a band-pass filter ($0.01$ Hz to $0.1$ Hz) to suppress physiological artifacts \cite{cai2019refined}. Finally, we reduced the dimensionality of the data by employing the standard $264$ ROIs template proposed by Power et al. \cite{Power2010} with a $5$ mm sphere radius. For the $264$ ROIs, we divide them into 12 functional network (FN) modules, including sensory/somatomotor network (SSN), cingulo-opercular task control network (CON), auditory network (AUD), default mode network (DMN), memory retrieval network (MRN), visual network (VN), fronto-parietal task control network (FPN), salience network (SN), subcortical network (SCN), ventral attention network (VAN), dorsal attention network (DAN) and cerebellum network (CERE).


Beyond imaging, all subjects underwent a $1$-hour long computerized neurocognitive battery (CNB) adapted from tasks applied in functional neuroimaging studies to evaluate a broad range of cognitive domains \cite{Moore2015}. For this work, we relied on performance scores (ratio of total correct responses) from the wide range achievement test (WRAT), which measures an individual's learning ability (reading, spelling, and mathematics) \cite{wilkinson2006}, and provides an estimate of intelligence quotient (IQ). To mitigate the influence of age over the final results, we excluded subjects whose ages were below 18 years old \cite{Xiao2018}. We then converted the WRAT scores to z-scores based upon each subject's raw score and the sample mean in order to provide a standard metric. We only kept subjects whose absolute z-score values were above 0.5, i.e., $|z| > 0.5$, to better compare and explain the brain cognitive differences. Finally, $193$ subjects were left for analysis and divided into 2 groups according to the WRAT scores (high WRAT group: $63$ subjects/low WRAT group: $130$ subjects). The detailed demographic information is summarized in TABLE \uppercase\expandafter{\romannumeral2}.

\begin{table}[h!]
\centering
\caption{Characteristics of the subjects in this study. Here SD denotes the standard deviation, M and F represent male and female respectively.}
\begin{tabular}{c|c|c|c}
\hline
Group &  Mean WRAT (SD) & Age Range& Sex Ratio (M:F) \tabularnewline
\hline
High &  $116.36$ $(7.81)$ & $18-21.75$ & $27:36$ \tabularnewline
Low &  $91.26$ $(9.74)$  & $18-22.58$ & $52:78$ \tabularnewline
\hline
\end{tabular}
\end{table}

\subsubsection{Results}

First of all, we calculated the Anderson–-Darling score as suggested in \cite{Ramsey2014} to ensure that the rsfMRI data meet our non-Gaussianity assumption. For each subject, we then applied the $\psi$-LiNGAM method, and obtained the corresponding weighted directed adjacency matrix $\bold{B}$ at the significance level $\alpha=0.05$ (with FDR correction). We extracted three network statistics, as suggested in \cite{zhangTMI2019}, i.e., density, transitivity, and global efficiency (GE), to reflect various aspects of the directed brain functional network. Density is the fraction of present connections to all possible connections, which also shows the mean network degree. The degree of a node is defined as the number of edges connected to the node. For a directed graph, the degree of a node is composed of two elements: in-degree and out-degree, which are named after the directions of the edges. Transitivity calculates the ratio between the number of triangles and the number of connected node triples. Transitivity is a measure of the brain functional segregation, i.e., the ability for specialized processing to occur within densely interconnected groups of brain regions \cite{Rubinov2010}. GE calculates the average inverse shortest path length, which indicates the ability to combine specialized information from distributed brain regions, also known as the functional integration ability. The network statistics are calculated through the Brain Connectivity Toolbox (https://sites.google.com/site/bctnet/). In TABLE \uppercase\expandafter{\romannumeral3}, we find that although the brain network density of the low WRAT group is much larger than that of the high WRAT group, there are no big differences between them according to transitivity and global efficiency.


\begin{table}[h!]
\centering
\caption{Network statistics}
\begin{tabular}{|c|c|c|c|}
\hline
Group & Density & Transitivity & Global Efficiency \tabularnewline
\hline
High & $0.1189$ & $0.0033$ & $0.0121$  \tabularnewline
Low & $0.2329$   & $0.0032$   & $0.0118$  \tabularnewline
\hline
\end{tabular}
\end{table}

\begin{figure}
\centering
\subfloat[Subfigure 1 list of figures text][high WRAT group]{\includegraphics[width = 1.6in]{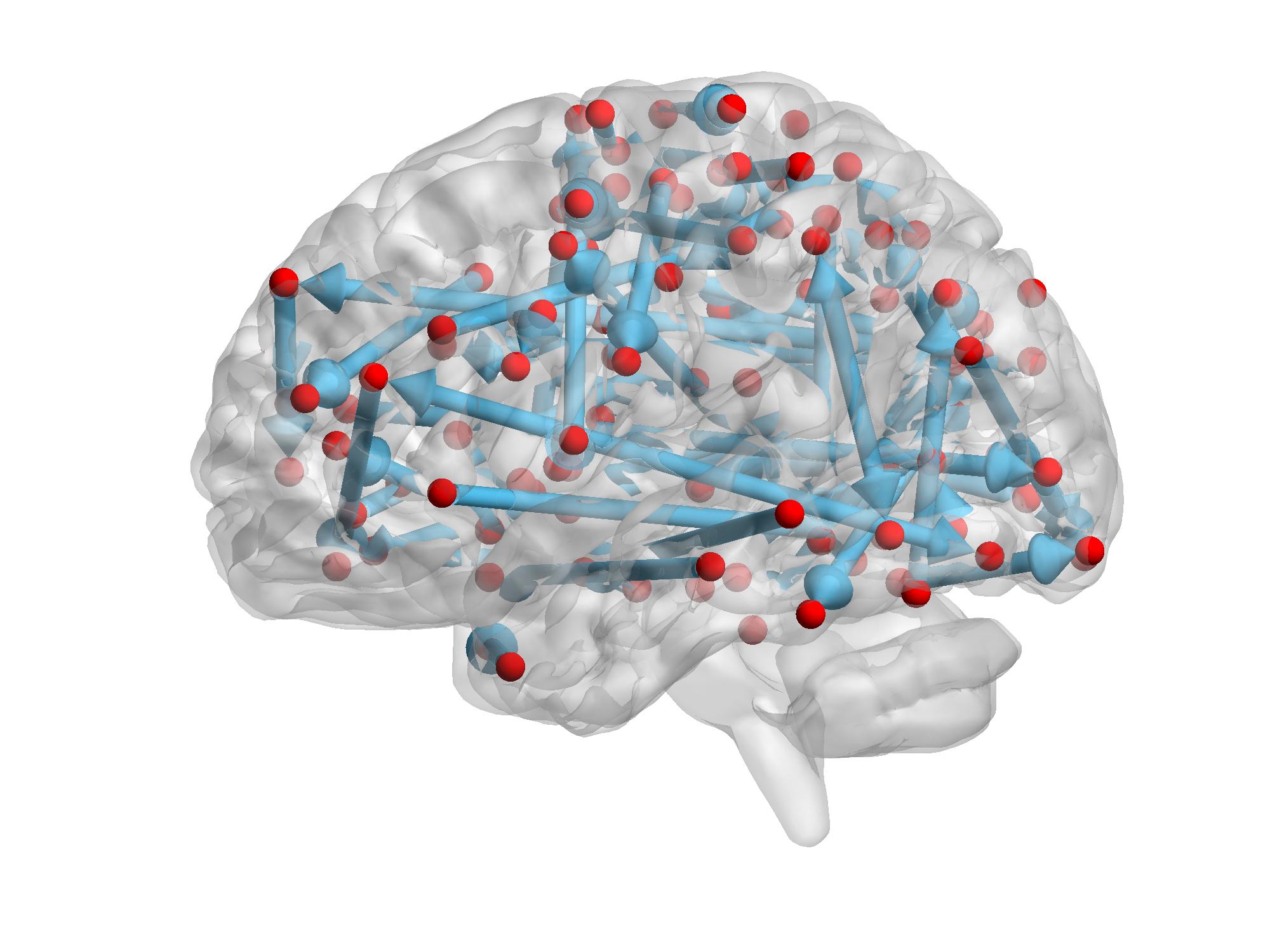}}
\qquad
\subfloat[Subfigure 1 list of figures text][low WRAT group]{\includegraphics[width = 1.6in]{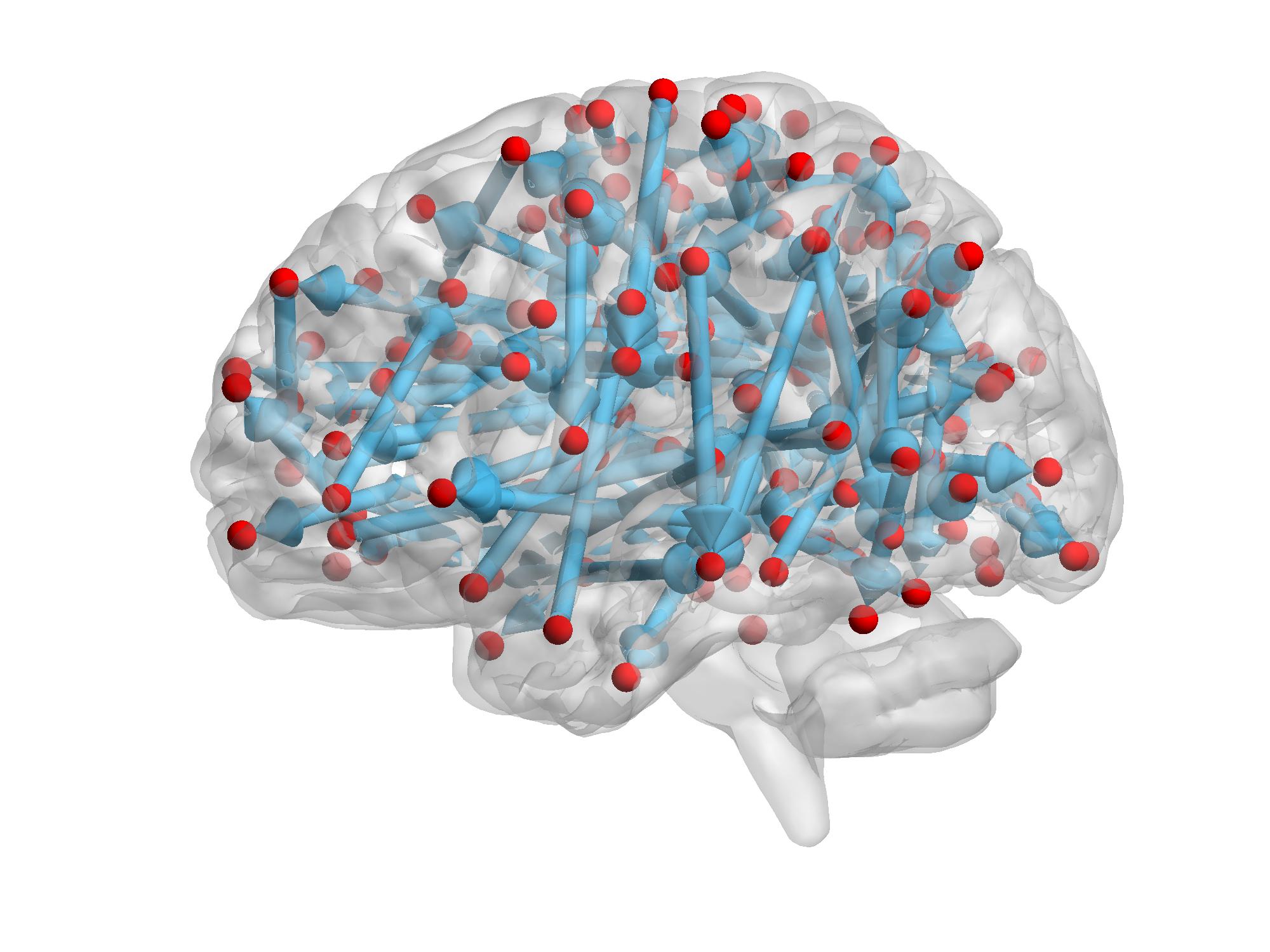}}
\qquad
\caption{Directed brain connectivity for each group, where the arrows indicate the causal flow.}\label{fig:3}
\end{figure}

To identify group structures, we conducted a one-sample t-test for each group to select the significant edges at significance level $\alpha = 0.05$ and set threshold of the weights $|b^*| = 0.1$ (i.e., the mean weights should be above $0.1$), which guarantee the results to be both statistically significant and numerically noticeable. Finally, $74$ and $140$ pairs of directed brain connections were identified for the high and low WRAT groups, respectively (see Fig. \ref{fig:3}). To gain more insights into the obtained networks, we analyzed the inferred hub nodes in three ways: in-hubs, out-hubs and sum-hubs. Here we define hubs as the nodes with degrees at least two standard deviation higher than the mean degrees (\cite{Jian2018}). The in-hubs (centers that receive information) and out-hubs (central nodes that convey out information) are selected through the in- and out- degrees, respectively, while the sum-hubs are the ones that do not belong to the in- and out- hubs but their sum degrees are high, i.e., transitional centers. As shown in TABLE \uppercase\expandafter{\romannumeral4}, there is one in-hub ROI located in the right inferior frontal gyrus, triangular (IFGT.R) and one out-hub ROI located in the right middle occipital gyrus (MOG.R) for the high WRAT group. For the low WRAT group, $4$ in-hub ROIs have been selected, which are located in the right middle frontal gyrus (MFG.R), right precentral gyrus (PREG.R), right inferior temporal gyrus (ITG.R), and right middle temporal gyrus (MTG.R); $1$ out-hub ROI is located at the right superior temporal gyrus (STG.R). There are no sum-hubs identified for both groups.

\begin{table}[h!]
\caption{ The hub ROIs from the identified directed network of each group, where AR represents the anatomical region.}
\centering
\begin{tabular}{c|c|c|c|c|c}
\hline
\multicolumn{6}{c}{High WRAT Group} \tabularnewline
\hline
Type &Index & MNI& AR & FN & Degree\tabularnewline
\hline
In-hub & 175 & $(48, 25, 27)$ & IFGT.R & FPN & $3$ \\
\hline
Out-hub & 157 & $(29, 77, 25)$ & MOG.R & VN & $16$ \\
\hline
\hline
\multicolumn{6}{c}{Low WRAT Group} \tabularnewline
\hline
Type &Index & MNI& AR & FN & Degree\tabularnewline
\hline
\multirow{4}{*}{In-hub} & 206 & $(31,33,26)$ & MFG.R & SN& $7$ \\
                        & 205 & $(40,0,47)$  & PRE.R & SN & $6$ \\
                        & 11  & $(55,-31,-17)$ & ITG.R & - & $5$ \\
                        & 84  & $(-58,-26,-15)$ & MTG.R & -& $5$ \\
\hline
Out-hub & 235 & $(54, -43, 22)$ & STG.R & VAN & $67$ \\
\hline
\end{tabular}
\end{table}

To further compare the differences between the high and low WRAT groups, we conducted a Welch two sample t-test at significance level $\alpha=0.05$ and Cohen's d effect size statistics \cite{cohen1988}. According to Cohen \cite{Cohen1992}, the difference is negligible when $|d| <0.2$. Thus, we chose $|d|>0.2$ and selected $1113$ significant edges that perform differently between the two groups. The numbers of edges (features) for various strength levels of difference are given in TABLE \uppercase\expandafter{\romannumeral5}. The hub ROIs discovered with $|d|>0.2$ have been summarized in TABLE \uppercase\expandafter{\romannumeral6}. Among them, $4$ in-hub ROIs are located at the right postcentral gyrus (POST.G), left precuneus (PQ.L), right precuneus (PQ.R) and left middle occipital gyrus (MOG.L); $2$ out-hub ROIs are located at the left rolandic operculum (RO.L) and right putamen (PUT.R); $3$ sum-hub ROIs are located at the right lingual gyrus (LING.R), left calcarine sulcus (CAL.L), and right thalamus (THA.R).

\begin{table}[h!]
\caption{The numbers of edges for different Cohen's d statistics thresholds.}
\centering
\begin{tabular}{|c|c|c|c|c|}
\hline
$|d|>$ & $0.5$ & $0.4$ & $0.3$ & $0.2$ \\
\hline
\# of edges & $2$  & $16$ & $141$ & $1113$\\
\hline
\end{tabular}
\end{table}


\begin{table}[h!]
\caption{The hub ROIs of the different connections selected from threshold $|d|>0.2$, where AR represents the anatomical region.}
\centering
\begin{tabular}{c|c|c|c|c|c}
\hline
Type &Index & MNI& AR & FN & Degree\tabularnewline
\hline
\multirow{5}{*}{In-hub} & 25 & $(50,-20,42)$ & POST.R & SSN& $20$ \\
                        & 90 & $(-11,-56,16)$  & PQ.L & DMN & $19$ \\
                        & 251& $(10,-62,61)$ & PQ.R & DAN & $18$ \\
                        & 147& $(-28,-79,19)$ & MOG.L & VN& $17$ \\
\hline
\multirow{2}{*}{Out-hub} & 70 & $(-55, -9, 12)$ & RO.L & AUD & $12$ \\
                         & 229& $(31, -14, 2)$ & PUT.R & SCN & $11$ \\
\hline
\multirow{3}{*}{Sum-hub} & 148& $(20,-66,2)$ & LING.R & VN & $20$ \\
                         & 146& $(-8,-81,7)$ & CAL.L & VN & $19$ \\
                         & 225& $(12,-17,8)$ & THA.R & SCN & $19$ \\
\hline

\end{tabular}
\end{table}

\subsubsection{Support Vector Machine based Classification}

With the available phenotype information (WRAT score high/low), we did classification based on our selected features (significantly different connections) as a validation to our discovery. Specifically, we applied linear support vector machine (SVM) with default parameter value $C=1$ \cite{scholkopf2002} and compared the results with $8$ various inputs, which are given as follows:

\begin{enumerate}
\item[a.] Pearson correlation (\cite{Cai2018}): Calculate the Pearson correlation matrix for each subject and use the upper triangular elements of the matrix as the input.
\item[b.] Partial correlation (\cite{Liang2015}): Apply the $\psi$-learning method to calculate the high-dimensional partial correlation matrix and use the upper triangular elements of the matrix as the input.
\item[c.] PC (\cite{pc2010}): Apply the PC-algorithm and acquire the adjacency matrix of the completed partially directed acyclic graph (CPDAG), which contains both undirected and directed edges. Use all the elements from the adjacency matrix as the input.
\item[d.] $\psi$-LiNGAM: Apply the proposed $\psi$-LiNGAM method to each subject and acquire the weighted adjacency matrix $\bold{B}$. Use all the elements from $\bold{B}$ as the input.
\item[e.] $|d|> 0.5$: Apply the proposed $\psi$-LiNGAM method to each subject first, then compare the differences between the two groups as described in \uppercase\expandafter{\romannumeral3}.B 2). Set $|d|> 0.5$ as the threshold for the feature selection (significantly different connections) and only choose the elements of the selected features from $\bold{B}$ as the input.
\item[f.] $|d|> 0.4$ : Same as e., except for setting $|d|> 0.4$ as the threshold.
\item[g.] $|d|> 0.3$ : Same as e., except for setting $|d|> 0.3$ as the threshold.
\item[h.] $|d|> 0.2$ : Same as e., except for setting $|d|> 0.2$ as the threshold.
\end{enumerate}

\begin{figure}
\centering
\includegraphics[width =3.3in]{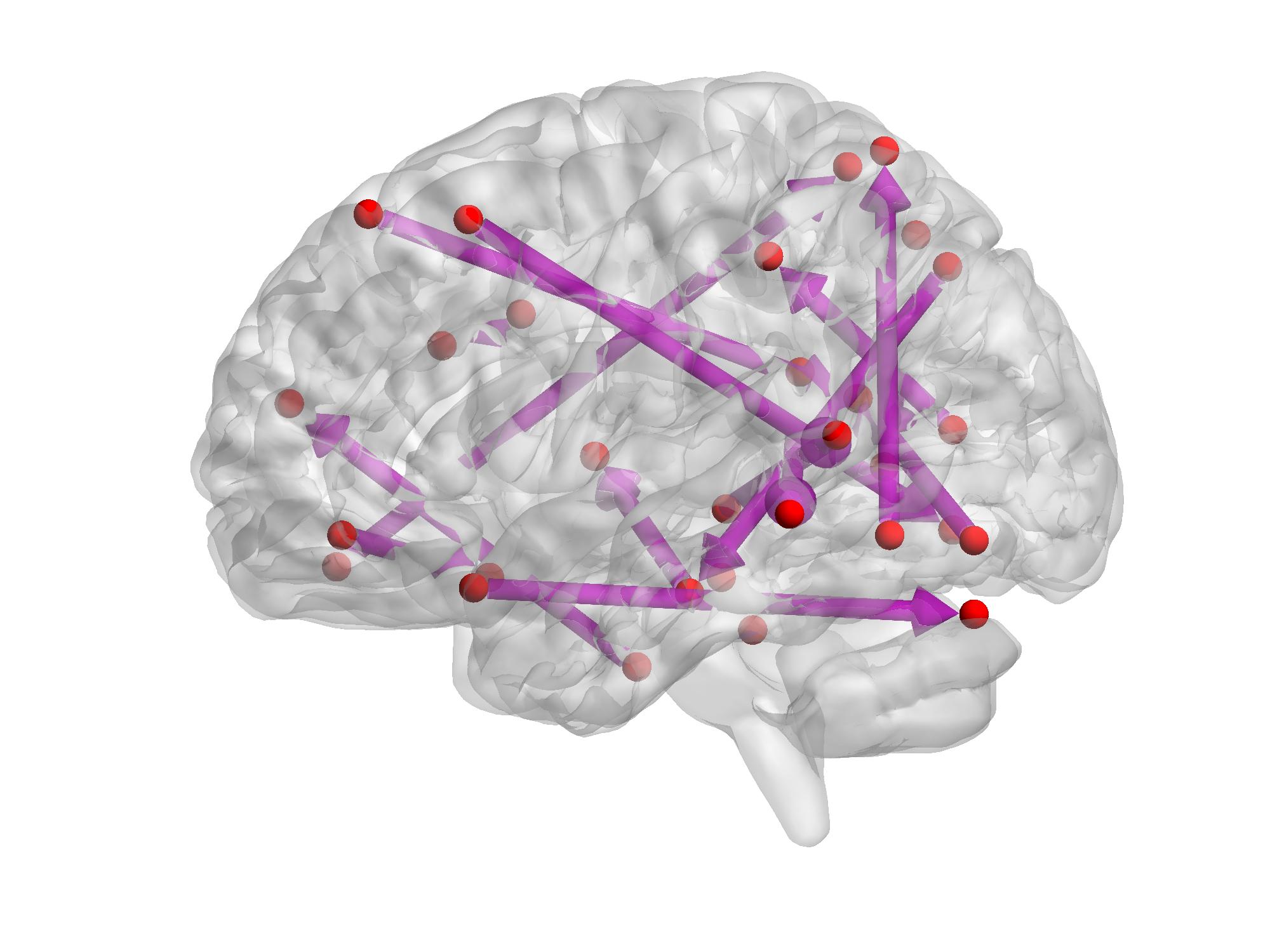}
\caption{Sagittal views of the $16$ causal connections, where the arrows indicate the causal flow.}\label{fig:4}
\end{figure}

All the algorithms were implemented in R. The $\psi$-learning method is available in the R package \emph{equSA}. The PC-algorithm was implemented through the R package \emph{pcalg}. The SVM algorithm was implemented through the R package \emph{e1071}. A 5-fold cross-validation (CV) procedure was implemented to evaluate the classification performance in all experiments. All subjects were randomly partitioned into 5 disjoint subsets with similar class distributions and size $= (39,39,39,39,37)$. Each subset in turn was used as the test set and the remaining 4 subsets were used to train the SVM classifier. The classifier accuracy was estimated by comparing against the ground-truth labels on the test set. We then averaged the 5 individual accuracy measures as one test result from the CV. The whole process was repeated 20 times to acquire the final mean accuracy rate (ACC) for each input. In TABLE \uppercase\expandafter{\romannumeral7}, we find that generally using the directed information (PC, $\psi$-LiNGAM) as inputs is better than using the undirected associations, however, the improvement is not very big. We then selected important features based on the different connections and set various thresholds to compare classification results. After feature selection from the $\psi$-LiNGAM result, the ACC has been improved. Especially, when we set $|d|>0.4$ and chose $16$ features as the input, the ACC has increased dramatically, which suggests that these $16$ features captured the major differences of brain connectivity between the high and low WRAT groups. The details of the $16$ causal connections are shown in Fig. \ref{fig:4} and TABLE \uppercase\expandafter{\romannumeral8}. When $|d|>0.2$, i.e., $1113$ features were selected, the ACC was the highest; it indicates that the selected features can explain the most cognitive variance between the two groups.

\begin{table}[h!]
\caption{ The mean classification results (ACC) by SVM with various inputs.}
\centering
\begin{tabular}{c|c|c|c|c}
\hline
Input & Pearson   & Partial   & PC        & $\psi$-LiNGAM \\
\hline
ACC   & $62.62\%$ & $60.77\%$ & $65.25\%$ & $67.31\%$\\
\hline
\hline
Input & $|d|>0.5$ & $|d|>0.4$  & $|d|>0.3$ & $|d|>0.2$  \\
\hline
ACC   & $70.87\%$ & $79.63\%$ & $82.37\%$ & $84.87\%$  \\
\hline
\end{tabular}
\end{table}

\begin{table*}[h!]
\caption{ The $16$ causal connections selected for $|d|>0.4$, where L and R represent the left and the right side of the brain, respectively.}
\centering
\setlength\tabcolsep{1.5pt}
\begin{tabular}{c|c|c|c|c|c|c|c|c}
\hline
\multicolumn{4}{c|}{ROI} & \multirow{2}{*}{Direction} & \multicolumn{4}{c}{ROI} \tabularnewline
\cline{1-4}\cline{6-9}
Index & MNI & AR (L/R) & FN & & Index & MNI & AR (L/R) & FN \\
\hline
\multirow{2}{*}{6} & \multirow{2}{*}{$(-21,-22,-20)$} & \multirow{2}{*}{Parahippocampus (L)}& \multirow{2}{*}{-} & $\longleftarrow$ & 134  & $(-7,-71,42)$ & Precuneus (L) & MRN \\
&&&& $\longrightarrow$ & 234 & $(9, -4, 6)$ & Thalamus (R) & SCN \\
\hline
7 & $(17,-28,-17)$ & Parahippocampus (R) & - & $\longrightarrow$ & 10 & $(52,-34,-27)$ & Inferior temporal gyrus (R) & - \\
\hline
98 & $(-10,39,52)$ & Superior frontal gyrus, & DMN & $\longrightarrow$ & \multirow{3}{*}{95} & \multirow{3}{*}{$(11,-54,17)$} & \multirow{3}{*}{Precuneus (R)} & \multirow{3}{*}{DMN}\\
&& medial (L)&&&&&&\\
154 & $(-47,-76,-10)$ & Inferior occipital lobe (L) & VN &$\longrightarrow$ &&&&\\
\hline
100 & $(-35,20,51)$ & Middle frontal gyrus (L) & DMN & $\longrightarrow$ & 151 & $(-15,-72,-8)$ & Lingual gyrus (L) & VN\\
\hline
109 & $(-3,44,-9)$ & Superior frontal gyrus, & DMN & $\longrightarrow$ & 85 & $(27,16,-17)$ & Insula (R) & - \\
&& medial orbital (L)&&&&&&\\
\hline
132 & $(-31,19,-19)$ & Inferior frontal gyrus,  & - & $\longrightarrow$ & 183 & $(-18,-76,24)$ & Cerebellum  & -\\
&& orbital (L)&&&&&crus 1 (L)&\\
\hline
145 & $(8,-72,11)$ & Calcarine (R) & VN & $\longrightarrow$ & 94 & $(-2,-37,44)$ & Middle cingulate gyrus (L) & DMN \\
\hline
180 & $(24,45,-15)$ & Superior frontal gyrus,  & FPN & $\longrightarrow$ & 13 & $(-7,-52,61)$ & Precuneus (L) & SSN \\
&& orbital (R)&&&&&& \\
\hline
186 & $(47,10,33)$ & Precentral gyrus (R) & FPN & $\longrightarrow$ & 175 & $(48,25,27)$ & Inferior frontal gyrus,  & FPN \\
&&&&&&& triangular (R)& \\
\hline
235 & $(54,-43,22)$ & Superior temporal gyrus (R) & VAN & $\longrightarrow$ & 120 & $(-68,-41,-5)$ & Middle temporal gyrus (L) & DMN \\
\hline
239 & $(51,-29,-4)$ & Middle temporal gyrus (R) & VAN & $\longrightarrow$ & 236 & $(-56,-50,10)$ & Middle temporal gyrus (L) & VAN\\
\hline
247 & $(33,-12,-34)$ & Fusiform gyrus (R) & - & $\longrightarrow$ & 105 & $(6,54,16)$ & Superior frontal gyrus,  & DMN\\
&&&&&&& medial (R)& \\
\hline
256 & $(22,-65,48)$ & Superior parietal gyrus (R) & DAN & $\longrightarrow$ & 257 & $(46,-59,4)$ & Middle temporal gyrus (R)& DAN\\
\hline
262 & $(-42,-60,-9)$& Inferior temporal gyrus (L) & DAN & $\longrightarrow$ & 263 & $(-17,-59,64)$ & Superior parietal gyrus (L)& DAN\\
\hline
\end{tabular}
\end{table*}

\subsubsection{Summary}

By applying the $\psi$-LiNGAM method, we obtained each subject's weighted adjacency matrix $\bold{B}$. From the mean network statistics extracted from $\bold{B}$ of each group, we found that although the mean densities of the two groups were significantly different (low $>$ high), there were no big differences in terms of transitivity and GE. In other words, the ability to combine and process specialized information was similar between the low and high WRAT groups; however, populations in the low WRAT group used more brain connections. The group-wise directed connectivity is shown in Fig. \ref{fig:3} and the detailed hub information is given in TABLE \uppercase\expandafter{\romannumeral4}. We found that both groups had an out-hub whose degree was relatively large, respectively. To be more specific, for the high WRAT group, the ROI located at the right middle occipital gyrus (MOG.R) has significant influences on other areas at rest, while the ROI located at the right superior temporal gyrus (STG.R) has sent extensive information to other ROIs. To compare the differences between groups, as suggested from the classification results, the selected features explained the most cognitive variance when setting $|d|>0.2$. We then identified $4$ in-hubs, $2$ out-hubs and $3$ sum-hubs (see TABLE \uppercase\expandafter{\romannumeral6}) from the different connections. Combining all the results above, $3$ functional modules (the VN, SN and SCN) have been mostly involved in the cognitive differences. Besides, we concluded that the left middle occipital gyrus (MOG.L), precuneus (PQ), and the right postcentral gyrus (POST.G) are influenced differently; left rolandic operculum (RO.L) and right putamen (PUT.R) have aberrant outgoing performance; the right lingual gyrus (LING.R), left calcarine sulcus (CAL.L), and right thalamus (THA.R), as the transitional areas, perform variously at different cognitive levels. TABLE \uppercase\expandafter{\romannumeral8} presents most important $16$ pairs of causal flows.

\section{Discussion}

In the procedure of identifying group brain connectivity patterns and feature selection (significantly different edges), we have used both significance tests (one-sample and two-sample t-tests) and effect size statistics (regression parameters and Cohen's d statistics). The reasons we did both are described as follows. Firstly, the sample sizes of the two groups are different, while the t-value and the corresponding p-value are dependent on the sample size, which makes the statistical results not comparable. The effect size is not dependent on the sample size, hence comes to be a rational choice. Secondly, effect size statistics provide a uniform standard, but the significance of the value stays unknown. In our study, we applied both to make sure that the results were statistically significant and numerically meaningful.

Some of our findings have been already validated in the literature. For instance, considerable prior studies have implicated that the precuneus (PQ), as a central node in the human brain, exhibits heightened connectivity within both DMN and task-positive networks while at rest \cite{leech2011}, and is important for supporting complex cognition and behavior \cite{utevsky2014}. The study in \cite{cavanna2006} showed that precuneus played an important role in integrating both internally and externally driven information. Kilory et al. \cite{kilroy2011} have found positive relationships between PQ and intelligence quotient (IQ). All the evidence agrees with our finding that PQ, as the in-side of the causal connections, is affected differently at rest for various cognitive ability (IQ) ranges. The postcentral gyrus (POST), otherwise known as the primary somatosensory cortex, has been reported to correlate with cognitive abilities \cite{brown2004, johnson2008}. Our results give a possible explanation that POST, as an in-hub of different causal connectivity in various cognitive groups, receives signals aberrantly from other ROIs. In this sense, the out-hub features (centers that convey out information differently) should attract more attention. Among them, Putamen (PUT) has been found to relate to many regions of the cerebral cortex, as well as the thalamus, claustrum, and others through various pathways. In many studies, it has become apparent that the putamen is involved in various types of learning, such as reinforcement and implicit learning \cite{yamada2004} and category learning \cite{ell2006}, and thus relates to IQ \cite{macdonald2014}. Our conclusion agrees with the previous studies and suggests that putamen is a center for sending out information that causes the cognitive variance. The superior temporal gyrus (STG), has been discovered to be an important structure in the pathway consisting of the amygdala and prefrontal cortex, which are all involved in cognition \cite{adolphs2003, bigler2007}. It is essential to the function of language in individuals who may have an impaired vocabulary, or are developing a sense of language. Several fMRI studies have suggested a link between insight based problem solving and activity in the right anterior superiortemporal gyrus (STG.R) \cite{jung2004}, which was identified as the out-hub ROI of the low WRAT group in our analysis. The thalamus has been generally believed to act as a relay station, relaying information between different subcortical areas and the cerebral cortex \cite{gazzaniga2013}, which strongly correlates with level of intelligence \cite{pergola2013}.

We have discovered some hub nodes in the occipital lobe, more precisely at the middle occipital gyrus (MOG), left rolandic operculum (RO.L) and the right lingual gyrus (LING.R). The lingual gyrus is a structure in the visual cortex,  which plays an important role in the identification and recognition of words \cite{mechelli2000}, and has been found to correlate with cognitive variance \cite{johnson2008}. The roles of MOG and RO.L in cognitive differences are new discoveries, which may suggest potential functional basis of cognition variances but still need further validations.

\section{Conclusion}

In this paper, we propose a $\psi$-learning incorporated linear non-Gaussian Acyclic model ($\psi$-LiNGAM) to study the casual interactions in human brain. As suggested by its name, the proposed method assumes non-Gaussianity of the data and believes that the non-Gaussian information can help to identify the direction of the edge. The main contributions of our work can be summarized as follows. First, the proposed method well integrates both the undirected graph and the directed acyclic graph, in the sense that we incorporate the undirected graph estimation as prior information into the direct LiNGAM model to perform DAG construction. Since the DAG is a subset of the undirected graph, the first step of prior screening can mitigate the irrelevant information but still maintain the true subset, which speeds up both numerical convergence and computation. Second, the simulation results in \uppercase\expandafter{\romannumeral3}. A show that the proposed method is stable with different settings and demonstrates improved performance in graph structure detection compared with $4$ other competing methods. Therefore, $\psi$-LiNGAM is a more solid statistical model for DAG estimation. Third, the proposed method has been applied to rs-fMRI data from PNC, to explain brain functions through directed FC differences. We have identified $1113$ significant causal connections that perform differently between the high and low WRAT groups. Three types of hub structures were found (see TABLE \uppercase\expandafter{\romannumeral6}). Several of our hub ROIs have already been validated to have correlations with the cognitive variance. Our findings can further illustrate the roles of the hub ROIs in the brain causal flow. Finally, the classification results in \uppercase\expandafter{\romannumeral3}.B 3) prove the reliability of our discoveries, which also indicate that it can be widely applied to feature selection tasks in many other neuroimaging studies.

Future work for this line of research includes the following. First, for more reliable biomedical discovery, more datasets are needed to cross validate our findings. Second, the PNC data contains nearly $900$ subjects of age from late childhood to young adulthood. In this work, we only used the subjects over $18$ years old. It would also be interesting to track functional causal brain connectivity development over various age periods, which requires a joint DAG estimation model such as \cite{wang2018} or time-series DAG model. We are currently working on the expansion of the $\psi$-LiNGAM method for causal inferences from multiple functional connectivity networks at different brain development stages, which will be reported elsewhere.

\section{Acknowledgment}
The work has been funded by NIH (R01GM109068, R01MH104680, R01MH107354, P20GM103472, 2R01EB005846, 1R01EB006841, R01MH121101, R01MH103220), and NSF (\#1539067).

\bibliographystyle{IEEEtran}
\bibliography{refer}

\end{document}